\documentclass[10pt]{article}
\usepackage[margin=1in]{geometry}
\usepackage{microtype}
\usepackage{booktabs}
\usepackage{enumitem}
\usepackage{amsmath,amssymb}
\usepackage[numbers,sort&compress]{natbib}
\usepackage{hyperref}
\hypersetup{colorlinks=true,allcolors=blue}
\usepackage{longtable}
\usepackage{graphicx}
\usepackage{tcolorbox}
\usepackage{listings}
\usepackage{color-edits}
\addauthor{vs}{red}
\addauthor{as}{blue}
\usepackage{booktabs,tabularx,makecell,array}
\newcolumntype{Y}{>{\raggedright\arraybackslash}X}

\title{CausalReasoningBenchmark: A Real-World Benchmark for Disentangled Evaluation of Causal Identification and Estimation \footnote{We used ChatGPT and Manus to edit parts of the manuscript.
For dataset curation, we used AI to draft summaries of the context in which each dataset was collected and descriptions of its columns.
We also used AI to generate reproducible estimation code for each causal query.
We used AI to review the questions and metadata over several passes to check for information that could reveal the answers.
The authors remain responsible for the dataset, code, and manuscript, including any errors.}\vspace{2pt}}
\author{
Ayush Sawarni, Jiyuan Tan, and Vasilis Syrgkanis\\[6pt]
Stanford University\\[2pt]
\texttt{\{ayushsaw,jiyuantan,vsyrgk\}@stanford.edu}
}
\date{}

\begin{document}
\maketitle

\begin{abstract}
\noindent
Many benchmarks for automated causal inference evaluate a system's performance based on a single numerical output, such as an Average Treatment Effect (ATE).
This approach conflates two distinct steps in causal analysis: \textbf{identification}---formulating a valid research design under stated assumptions---and \textbf{estimation}---implementing that design numerically on finite data.
We introduce \textbf{CausalReasoningBenchmark}, a benchmark of 174 queries across 139 real-world datasets, curated from 73 peer-reviewed research papers and three widely-used causal-inference textbooks.
For each query a system must produce (i)~a structured identification specification that names the strategy, the treatment, outcome, and control variables, and all design-specific elements, and (ii)~a point estimate with a standard error.
By scoring these two components separately, our benchmark enables granular diagnosis: it distinguishes failures in causal reasoning from errors in numerical execution.
Baseline results with a state-of-the-art LLM show that, while the model correctly identifies the high-level strategy in 81\% of cases, full identification-specification correctness drops to only 38\%, revealing that the bottleneck lies in the nuanced details of research design rather than in computation.
CausalReasoningBenchmark \footnote{\url{https://huggingface.co/datasets/syrgkanislab/CausalReasoningBenchmark}} is publicly available on Hugging Face and is designed to foster the development of more robust automated causal-inference systems.
\end{abstract}

\vspace{-6pt}
\section{Introduction}
\label{sec:intro}

LLMs and LLM-based agents are increasingly used for causal reasoning over observational data. However, the evaluation of these systems often falls short of the rigor required for real-world applications.
A common practice is to assess a model's performance based solely on a single numerical output---typically an effect estimate such as the Average Treatment Effect (ATE).
This single-output evaluation is limited because it conflates two distinct steps that are central to any empirical causal analysis.

The first step is \textbf{identification}: a conceptual exercise in which the analyst determines whether a causal quantity of interest is recoverable from the available data, given a set of assumptions about the data-generating process.
This requires specifying a valid research design---often called an \emph{identification strategy}---such as an Instrumental Variable (IV) design, a Regression Discontinuity Design (RDD), or a Difference-in-Differences (DiD) design, and defining all of its necessary components (e.g., the instrument, the running variable and cutoff, or the time and group indices).
The second step is \textbf{estimation}: a numerical exercise in which the identified strategy is implemented on a finite data sample to compute a point estimate of the causal effect and to quantify the uncertainty around that estimate.

Existing benchmarks typically collapse these two steps into a single score, making it impossible to diagnose the source of errors.
Did the model fail because it chose an invalid identification strategy, or did it implement a valid strategy incorrectly?
Furthermore, many benchmarks rely on synthetic or simplified data, which may not reflect the complexities of real-world empirical research: missing data, ambiguous variable definitions, and study-specific design choices.

To address these gaps, we introduce \textbf{CausalReasoningBenchmark}, a benchmark for evaluating automated causal reasoning systems.
Our main contributions are:
\begin{enumerate}[leftmargin=*]
    \item \textbf{A curated real-world benchmark.} We curate 174 queries over 139 unique datasets from 73 peer-reviewed research papers and three causal-inference textbooks.
    \item \textbf{A structured identification schema.}  The benchmark requires agents to specify not only the design family, but also the estimand, treatment, outcome, controls, and design-specific fields required for IV, RDD, DiD, conditional-exogeneity, and RCT designs.
    \item \textbf{Disentangled evaluation of identification and estimation} We provide gold identification specifications, reference estimation scripts, and an evaluator that separately scores research-design correctness and numerical estimation accuracy.
    \item \textbf{Standardized estimation scripts.} We provide gold-standard estimation code for every query, allowing failures in identification to be isolated from failures in implementation.
\end{enumerate}

The rest of this paper is organized as follows.
Section~\ref{sec:related_work} reviews related benchmarks and agents.
Section~\ref{sec:why_separate} motivates the separation of identification and estimation.
Section~\ref{sec:dataset} describes the CausalReasoningBenchmark dataset.
Section~\ref{sec:strategies} provides a formal description of the identification strategies covered.
Section~\ref{sec:io} defines the evaluation task and metrics.
Section~\ref{sec:llm-baseline} presents baseline results, including a qualitative analysis of identification errors.
Section~\ref{sec:case} walks through a concrete example.
Section~\ref{sec:hosting} discusses hosting and maintenance.
Section~\ref{sec:limitations} addresses limitations and future work.
Section~\ref{sec:conclusion} concludes.

%% ===== RELATED WORK =====
\section{Related Work}
\label{sec:related_work}

We situate CausalReasoningBenchmark relative to two lines of work: benchmarks for causal reasoning and LLM-based causal-inference agents.

\paragraph{Benchmarks for Causal Reasoning.}
\citet{liu2024qrdata} introduce \textbf{QRData}, a benchmark of quantitative reasoning tasks over spreadsheet-style data, including some causal estimation problems.
While QRData tests a broad range of data-analysis skills, it does not focus on the specific identification strategies used in observational studies and does not evaluate identification separately from estimation.
\citet{zhou2024causalbench} present \textbf{CausalBench}, which evaluates correlation, causal-skeleton, and causality-identification tasks over causal networks.
CausalBench is valuable for testing general causal reasoning, but it does not require systems to produce a full identification specification for quasi-experimental designs.
\citet{lee2025causal} build a benchmark by extracting validated, but not quantitative, cause--effect relations from economics and policy papers.
Their dataset includes common designs such as IV, DiD, and RDD, but---like the others---it evaluates qualitative causal relations rather than effect estimates, making it difficult to distinguish between identification and estimation errors.

\paragraph{LLM-Based Causal-Inference Agents.}
Several agent-based systems have been developed to automate parts of the causal inference workflow.
\textbf{CATE-B}~\citep{berrevoets2025cateb} is an LLM co-pilot that constructs directed acyclic graphs (DAGs), selects adjustment sets, and suggests estimators.
\textbf{ORCA}~\citep{chung2025orca} connects LLMs to causal inference libraries (e.g., \texttt{DoWhy}) to load data, fit models, and summarize results.
In the biomedical domain, \textbf{MRAgent}~\citep{xu2025mragent} automates Mendelian randomization analyses using GWAS datasets.
These systems are typically evaluated on internal or synthetic tasks.
A separate line of work focuses on causal reasoning over graphical models. \citet{jin2023cladder} introduce \textbf{CLadder}, a benchmark for formal causal reasoning on synthetic graphs, testing aspects like identifying confounding bias. \citet{corr2cause2023} propose \textbf{Corr2Cause}, which tasks models with inferring causal relationships from correlational statements. \citet{sheth2025causalgraph2llm} present \textbf{CausalGraph2LLM}, a large-scale benchmark with over 700k queries on diverse causal graphs.  While these benchmarks are crucial for evaluating graph-based and counterfactual reasoning, they do not focus on the quasi-experimental designs common in applied empirical research.

CausalReasoningBenchmark provides a challenging, external evaluation suite derived from peer-reviewed research, with a focus on disentangling identification from estimation.

Table~\ref{tab:related-comparison} summarizes the key differences between CausalReasoningBenchmark and the most closely related benchmarks.

\begin{table}[t]
\centering
\footnotesize
\setlength{\tabcolsep}{3.5pt}
\renewcommand{\arraystretch}{1.1}
\caption{Comparison of CausalReasoningBenchmark with related benchmarks. ``ID eval'' indicates whether identification is evaluated separately from estimation. ``Real data'' indicates whether the benchmark uses real-world (non-synthetic) datasets. ``Quant. eval'' means evaluating causal effect estimation from data (e.g., ATE / ATT / LATE / CATE). ``Design-specific'' indicates whether the benchmark requires specification of design-specific elements (e.g., instruments, running variables).}
\label{tab:related-comparison}
\begin{tabularx}{\linewidth}{@{}Y c c c c c >{\raggedright\arraybackslash}p{2.35cm}@{}}
	\toprule
	\textbf{Benchmark} &
	\makecell{\textbf{\#}\\\textbf{Queries}} &
	\makecell{\textbf{Real}\\\textbf{data}} &
	\makecell{\textbf{ID}\\\textbf{eval}} &
	\makecell{\textbf{Quant.}\\\textbf{eval}} &
	\makecell{\textbf{Design-}\\\textbf{specific}} &
	\makecell{\textbf{Designs}\\\textbf{covered}} \\
	\midrule
	QRData \citep{liu2024qrdata} & 411 & \checkmark &  & Partial &  & Mixed \\
	CausalBench \citep{zhou2024causalbench} & 495 & Partial &  &  &  & Mixed \\
	CLadder \citep{jin2023cladder} & 10k &  & \checkmark &  &  & Graph-based \\
	Corr2Cause \citep{corr2cause2023} & 413k &  & \checkmark &  &  & Graph-based \\
	CausalGraph2LLM \citep{sheth2025causalgraph2llm} & 700k+ &  & \checkmark &  &  & Graph-based \\
	\textbf{CausalReasoningBenchmark} & \textbf{174} & \checkmark & \checkmark & \checkmark & \checkmark & IV, RDD, DiD, CE, RCT \\
	\bottomrule
\end{tabularx}
\end{table}

%% ===== WHY SEPARATE =====
\section{Why Separate Identification from Estimation?}
\label{sec:why_separate}

Separating identification from estimation mirrors the structure of empirical causal analysis. In applied research, identification is where the core intellectual contribution resides: it requires understanding the data-generating process, articulating the assumptions under which a causal quantity is recoverable, and specifying all the components of a valid research design.
Estimation, by contrast, is largely a technical exercise: given a correctly specified design, the choice of estimator (e.g., two-stage least squares for IV, local polynomial regression for RDD, or a two-way fixed-effects model for DiD) is often well-understood and can even be automated.

This distinction has practical consequences for evaluation.
Consider a model that correctly identifies an IV design and names the right instrument, treatment, and outcome, but makes a coding error in the two-stage least squares implementation.
Under a single-score evaluation, this model would receive the same score as one that misidentifies the entire research design.
By scoring identification and estimation separately, CausalReasoningBenchmark can distinguish between these two very different failure modes.

Moreover, the identification specification itself is a rich, structured object that can be evaluated along multiple dimensions.
For example, a model might correctly identify the strategy (IV) and the instrument, but fail to exclude a post-treatment variable from the control set, a critical error that would bias the estimate.
Our evaluation framework captures this level of detail, as described in Section~\ref{sec:io}.

%% ===== DATASET =====
\section{The CausalReasoningBenchmark Dataset}
\label{sec:dataset}

CausalReasoningBenchmark is designed to evaluate an agent's ability to correctly specify and execute a causal analysis.
It focuses on the canonical research designs used in observational studies: Instrumental Variables (IV), Regression Discontinuity (RD), Difference-in-Differences (DiD), Conditional Exogeneity (selection on observables), and Randomized Controlled Trials (RCT).
The benchmark consists of 174 queries over 139 datasets (see Tables~\ref{tab:cb-source-group}--\ref{tab:cb-strategy-by-source}).
Each query includes:
\begin{itemize}
    \item A natural-language causal question.
    \item A dataset in CSV format.
    \item A metadata file describing the variables and providing study context.
    \item A gold-standard solution, including a detailed identification specification (as a JSON object) and a reference estimation script (in Python or R).
\end{itemize}

The dataset is sourced from two main categories, described below.

\paragraph{Research Papers.}
We curated 120 queries from 73 papers, drawing from three large-scale reanalysis studies in political science:
\begin{itemize}
  \item \textbf{IV}: \citet{lal2024much} provide a replication of 67 instrumental variable studies.
  \item \textbf{RDD}: \citet{stommes2023reliability} examine 45 regression discontinuity studies and obtain replication data for 36.
  \item \textbf{DiD}: \citet{chiu2026causal} conduct a reanalysis of 49 panel data studies.
\end{itemize}
We selected cases from these corpora where the original paper presented a clear and defensible identification strategy, and where the documentation was sufficient to reconstruct the analysis.
The research-paper subset spans three top political science journals, providing a diverse set of real-world causal problems.

\paragraph{Textbook and Instructional Collections.}
To include classic and pedagogical examples, we added 54 queries from three popular causal inference textbooks:
\begin{itemize}
    \item \emph{Causal Inference: The Mixtape}~\cite{cunningham2021mixtape}
    \item \emph{The Effect: An Introduction to Research Design and Causality}~\cite{huntingtonklein2022effect}
    \item \emph{Causal Inference: What If}~\cite{hernanrobins2020whatif}
\end{itemize}
Several of these examples also appeared in  \emph{causaldata} R-package, \cite{huntington1causaldata} and \emph{QR Dataset} \cite{liu2024qrdata} which served as a starting point for us. The textbook subset complements the research-paper subset by providing well-documented examples with clear pedagogical intent, and by adding coverage of Conditional Exogeneity designs (40 queries) that are not represented in the research-paper subset (Table~\ref{tab:cb-strategy-by-source}).

\paragraph{Dataset Composition.}
Tables~\ref{tab:cb-source-group}--\ref{tab:cb-strategy-by-source} provide a detailed breakdown of the benchmark's composition.
Table~\ref{tab:cb-source-group} shows the split between research papers and textbooks.
Table~\ref{tab:cb-strategy} shows the distribution across identification strategies: DiD is the most common (67 queries), followed by RDD (44), Conditional Exogeneity (40), IV (22), and RCT (1).
Table~\ref{tab:cb-strategy-by-source} reveals that the research-paper subset is dominated by DiD, RDD, and IV, while the textbook subset provides the bulk of the Conditional Exogeneity examples.
The research-paper subset spans three top political science journals---\emph{The Journal of Politics}, \emph{American Journal of Political Science}, and \emph{American Political Science Review}---which account for the vast majority of the research-paper queries.

\begin{table}[t]
\centering
\caption{CausalReasoningBenchmark queries and datasets by source group.}
\label{tab:cb-source-group}
\begin{tabular}{lrr}
\toprule
Source group & \#queries & \#datasets \\
\midrule
Research papers & 120 & 85 \\
Textbook & 54 & 54 \\
\midrule
Total & 174 & 139 \\
\bottomrule
\end{tabular}
\end{table}

\begin{table}[t]
\centering
\caption{CausalReasoningBenchmark composition by identification strategy.}
\label{tab:cb-strategy}
\begin{tabular}{lrr}
\toprule
Identification strategy & \#queries & \#datasets \\
\midrule
Difference-in-Differences & 67 & 37 \\
Regression Discontinuity & 44 & 39 \\
Instrumental Variable & 22 & 22 \\
Conditional Exogeneity & 40 & 40 \\
RCT & 1 & 1 \\
\bottomrule
\end{tabular}
\end{table}

\begin{table}[t]
\centering
\caption{Query counts by source group and identification strategy.}
\label{tab:cb-strategy-by-source}
\small
\begin{tabular}{lrrrrr}
\toprule
Source group & DiD & RDD & IV & Cond. Exog. & RCT \\
\midrule
Research papers & 62 & 39 & 19 & 0 & 0 \\
Textbook & 5 & 5 & 3 & 40 & 1 \\
\bottomrule
\end{tabular}
\end{table}

% The textbook subset complements the research-paper subset by providing well-documented examples with clear pedagogical intent, and by adding coverage of Conditional Exogeneity designs (40 queries) that are not represented in the research-paper subset (Table~\ref{tab:cb-strategy-by-source}).

% \paragraph{Textbook:}
% To complement paper-sourced instances with standard teaching cases, we include 54 queries drawn from the following textbooks:
% \begin{itemize}
%     \item \emph{causaldata}: \cite{huntington1causaldata}
%     \item \emph{QR Dataset}:  Questions of quantitative reasoning with data \cite{liu2024qrdata}
% \end{itemize}

%% ===== IDENTIFICATION STRATEGIES =====
\section{Identification Strategies}
\label{sec:strategies}

A key feature of CausalReasoningBenchmark is that it requires systems to produce a \emph{structured identification specification} for each query.
This section provides a brief formal description of each identification strategy covered by the benchmark, along with the specific fields that the system must specify.

\subsection{Instrumental Variables (IV)}

An instrumental variable design exploits an exogenous source of variation (the \emph{instrument}, $Z$) that affects the treatment ($D$) but has no direct effect on the outcome ($Y$) except through $D$. The key assumptions are: (i) \emph{relevance}: $Z$ is correlated with $D$; (ii) the \emph{exclusion restriction}: $Z$ affects $Y$ only through $D$; and (iii) \emph{independence}: $Z$ is independent of the potential outcomes and potential treatment statuses.

In many applications, these assumptions may only be plausible after conditioning on a set of pre-treatment covariates $\mathbf{X}$. The IV assumptions are thus relaxed to hold conditionally: (i) \emph{conditional relevance}: $\text{Cov}(D, Z \mid \mathbf{X}) \neq 0$; (ii) \emph{conditional exclusion}: holding treatment fixed, the instrument has no effect on the outcome, $Y(d,z)=Y(d)$; and (iii) \emph{conditional independence}: $Z$ is independent of all potential outcomes and potential treatment statuses conditional on $\mathbf{X}$ ($Z \perp\!\!\!\perp \{Y(d), D(z)\}_{d,z} \mid \mathbf{X}$). For binary treatment and instrument, these conditions, together with monotonicity (no defiers), identify the LATE with appropriate covariate adjustment. Under the corresponding unconditional assumptions, the LATE for compliers is identified as:
\begin{equation}
\text{LATE} = \frac{\text{Cov}(Y, Z)}{\text{Cov}(D, Z)}.
\end{equation}

\noindent\textbf{Required fields:} \texttt{strategy} = \texttt{Instrumental Variable}; \texttt{instrument} (column name(s) of the instrument); \texttt{is\_encouragement\_design} (whether the instrument is a randomized binary encouragement); \texttt{treatments}; \texttt{outcomes}; \texttt{controls}; \texttt{causal\_quantity} (typically \texttt{LATE}).

\subsection{Regression Discontinuity (RDD)}

A regression discontinuity design exploits a known threshold (the \emph{cutoff}) on a continuous \emph{running variable} ($X$) that determines treatment assignment.
Units just above and just below the cutoff are assumed to be comparable, so, when treatment switches from 0 to 1 at the cutoff, the sharp-RD effect is identified as the discontinuity in the conditional expectation of the outcome at the cutoff:
\begin{equation}
\tau_{\text{RDD}} = \lim_{x \downarrow c} E[Y \mid X = x] - \lim_{x \uparrow c} E[Y \mid X = x],
\end{equation}
where $c$ is the cutoff value.
In a \emph{sharp} design, treatment is a deterministic function of the running variable; in a \emph{fuzzy} design, the probability of treatment changes discontinuously at the cutoff, and the design is analogous to an IV with the threshold indicator as the instrument.
Under the usual continuity, exclusion, and monotonicity assumptions, the fuzzy-RD effect for compliers at the cutoff is the outcome discontinuity divided by the nonzero discontinuity in treatment probability.

\noindent\textbf{Required fields:} \texttt{strategy} = \texttt{Regression Discontinuity}; \texttt{running\_variable} (column name); \texttt{cutoff} (numeric threshold); \texttt{treatments}; \texttt{outcomes}; \texttt{controls}; \texttt{causal\_quantity}.

\subsection{Difference-in-Differences (DiD)}

A difference-in-differences design compares the change in outcomes over time between a treated group and a control group. The key assumption is \emph{parallel trends}: in the absence of treatment, the average outcomes for the treated and control groups would have followed parallel paths over time.

This assumption can be relaxed to a \emph{conditional parallel trends} assumption, which posits that parallel trends hold after conditioning on a set of pre-treatment covariates $\mathbf{X}$. This allows the baseline trends to differ, as long as they are parallel within strata defined by $\mathbf{X}$. Formally, the assumption is $E[Y(0)_{\text{post}} - Y(0)_{\text{pre}} \mid D=1, \mathbf{X}] = E[Y(0)_{\text{post}} - Y(0)_{\text{pre}} \mid D=0, \mathbf{X}]$. Under the unconditional assumption, the Average Treatment Effect on the Treated (ATT) is identified as:
\begin{equation}
\tau_{\text{DiD}} = \bigl(E[Y_{\text{post}} \mid D=1] - E[Y_{\text{pre}} \mid D=1]\bigr) - \bigl(E[Y_{\text{post}} \mid D=0] - E[Y_{\text{pre}} \mid D=0]\bigr).
\end{equation}

\noindent\textbf{Required fields:} \texttt{strategy} = \texttt{Difference-in-Differences}; \texttt{time\_variable} (column name of the time index); \texttt{group\_variable} (column name of the unit/group identifier); \texttt{treatments}; \texttt{outcomes}; \texttt{controls}; \texttt{causal\_quantity} (typically \texttt{ATT}).

\subsection{Conditional Exogeneity (Selection on Observables)}

Under conditional exogeneity, treatment assignment is assumed to be independent of potential outcomes after conditioning on a set of observed covariates $\mathbf{X}$:
\begin{equation}
(Y(0), Y(1)) \perp\!\!\!\perp D \mid \mathbf{X}.
\end{equation}
Together with consistency and the appropriate overlap condition, this assumption, also known as \emph{unconfoundedness} or \emph{selection on observables}, allows identification of the ATE (or ATT) via regression adjustment, inverse probability weighting, or matching.

\noindent\textbf{Required fields:} \texttt{strategy} = \texttt{Conditional Exogeneity}; \texttt{treatments}; \texttt{outcomes}; \texttt{controls} (the conditioning set $\mathbf{X}$, which must include a minimal sufficient adjustment set); \texttt{causal\_quantity}.

\subsection{Randomized Controlled Trials (RCT)}

In a randomized controlled trial, treatment is assigned randomly, so identification is straightforward: the ATE is simply the difference in mean outcomes between the treated and control groups.
While RCTs are the gold standard for causal inference, they are included in CausalReasoningBenchmark primarily for completeness (1 query).

\noindent\textbf{Required fields:} \texttt{strategy} = \texttt{RCT}; \texttt{treatments}; \texttt{outcomes}; \texttt{controls}; \texttt{causal\_quantity}.

%% ===== EVALUATION TASK =====
\section{Evaluation Task and Metrics}
\label{sec:io}

\subsection{Task Definition}

For each query in the benchmark, an agent is provided with the following inputs:
\begin{itemize}
  \item \textbf{Question}: A causal query in natural language.
  \item \textbf{Dataset}: A CSV file containing the data.
  \item \textbf{Metadata}: A text file with column descriptions and study context.
\end{itemize}

\noindent The agent must produce two outputs:
\begin{enumerate}
  \item \textbf{Identification Specification}: A structured JSON object that adheres to the schema described in Section~\ref{sec:strategies}, detailing the chosen identification strategy and all its components.
  \item \textbf{Estimation Output}: A point estimate of the causal effect (\texttt{effect\_estimate}) and its standard error (\texttt{standard\_error}).
\end{enumerate}

\subsection{Identification Metrics}
\label{sec:ident_metrics}

We evaluate the identification specification by comparing it field-by-field with the gold standard.
The evaluator checks the following conditions:

\begin{itemize}[leftmargin=*]
  \item \textbf{Strategy}: Exact match of the identification strategy label (e.g., \texttt{Instrumental Variable}).
  \item \textbf{Causal quantity}: Exact match of the estimand label (e.g., \texttt{ATE}, \texttt{LATE}).
  \item \textbf{Treatments and outcomes}: Exact set match of the variable names specified by the agent against the gold standard.
  \item \textbf{Controls}: We check two conditions: (1)~the agent's specified controls must be a superset of the gold-standard \emph{minimal sufficient adjustment set}---the smallest set of pre-treatment covariates needed for identification (e.g., to satisfy conditional exogeneity or conditional parallel trends); and (2)~the agent's controls must not include any variables from the gold-standard \emph{bad controls} list, which includes post-treatment variables, mediators, and colliders whose inclusion would bias the estimate.
  \item \textbf{Strategy-specific fields}: Depending on the strategy, we require correct specification of all compulsory fields: for IV, the \texttt{instrument} list and \texttt{is\_encouragement\_design} flag; for RDD, the \texttt{running\_variable} and \texttt{cutoff}; for DiD, the \texttt{time\_variable} and \texttt{group\_variable}.
  \item \textbf{Overall identification correctness}: A binary indicator that is \texttt{true} only if \emph{all} of the above checks pass. This is the strictest metric and captures whether the model has fully specified a valid research design.
\end{itemize}

\subsection{Estimation Metrics}
\label{sec:est_metrics}

Given a predicted effect $\widehat{\tau}$ and standard error $\widehat{\mathrm{SE}}$, and gold-standard values $\tau^\star$ and $\mathrm{SE}^\star$, we form 95\% Wald confidence intervals $\mathrm{CI}_{\text{pred}} = [\widehat{\tau} - 1.96\,\widehat{\mathrm{SE}},\; \widehat{\tau} + 1.96\,\widehat{\mathrm{SE}}]$ and $\mathrm{CI}_{\text{gold}} = [\tau^\star - 1.96\,\mathrm{SE}^\star,\; \tau^\star + 1.96\,\mathrm{SE}^\star]$, and compute:

\begin{itemize}[leftmargin=*]
  \item \textbf{Point-estimate error}: Absolute error $|\widehat{\tau}-\tau^\star|$, signed error $\widehat{\tau}-\tau^\star$, and (when $\tau^\star \neq 0$) relative absolute error $\frac{|\widehat{\tau}-\tau^\star|}{|\tau^\star|}\times 100\%$.
  \item \textbf{Estimate within gold CI}: Whether $\widehat{\tau} \in \mathrm{CI}_{\text{gold}}$.
  \item \textbf{Null-hypothesis agreement}: Whether both intervals lead to the same reject/fail-to-reject decision for $H_0: \tau = 0$.
  \item \textbf{Opposite-direction flag}: Whether both intervals reject $H_0$ but imply opposite effect signs---a particularly dangerous type of error.
  \item \textbf{Interval overlap (Jaccard)}: We measure the overlap between the two confidence intervals using the Jaccard index:
  \begin{equation}
    J(\mathrm{CI}_{\text{pred}},\,\mathrm{CI}_{\text{gold}}) = \frac{|\mathrm{CI}_{\text{pred}} \cap \mathrm{CI}_{\text{gold}}|}{|\mathrm{CI}_{\text{pred}} \cup \mathrm{CI}_{\text{gold}}|},
  \end{equation}
  which equals 0 when the intervals are disjoint and 1 when they coincide.
  \item \textbf{CI Overlap}: A binary indicator of whether the predicted and gold-standard confidence intervals overlap, i.e., $\mathbf{1}[\mathrm{CI}_{\text{pred}} \cap \mathrm{CI}_{\text{gold}} \neq \emptyset]$.
  \item \textbf{Standard-error gap}: $|\widehat{\mathrm{SE}} - \mathrm{SE}^\star|$ and the relative gap $\frac{|\widehat{\mathrm{SE}} - \mathrm{SE}^\star|}{\mathrm{SE}^\star}$.
\end{itemize}

\paragraph{Auto-rescaling.}
A common source of spurious estimation error is a unit mismatch (e.g., an effect reported in percentage points vs. proportions). For example, if the gold-standard effect is 0.05 (a 5 percentage point increase) and the model predicts 5.0, a naive error calculation would be enormous.
To mitigate this, the evaluator can optionally rescale the predicted effect and standard error by a multiplicative factor from a fixed candidate set (e.g., $\{0.01, 0.1, 10, 100\}$). The evaluator selects the factor that minimizes the absolute error. In the example above, multiplying the prediction of 5.0 by 0.01 yields 0.05, which perfectly matches the gold standard. The evaluation would proceed with this rescaled value.
This ensures that trivial unit-conversion errors do not dominate the estimation metrics.

%% ===== LLM BASELINE =====
\section{LLM Baseline Evaluation}
\label{sec:llm-baseline}

To demonstrate the utility of our benchmark, we evaluated a simple
LLM-agent baseline. The baseline was run with \texttt{gpt-5.3} with reasoning. For each query, the runner supplied
the model with the natural-language causal question, an inline metadata
description, and the corresponding CSV dataset. For the primary execution path,
the CSV file was uploaded as a file attachment and made available to the API's
Python code-interpreter environment. The model was instructed to inspect the
data, write and execute Python code for the estimate and standard error, and
return a single JSON object containing both a schema-conformant identification
specification and the estimation code. The returned identification payload was
validated against our output schema before being scored. The full prompt template
is shown below.

\begin{tcolorbox}[colback=gray!5!white,colframe=gray!75!black,title=LLM prompt template]
\begin{lstlisting}[basicstyle=\ttfamily\scriptsize,breaklines=true]
SYSTEM:
You are a meticulous causal inference research assistant. You can read
structured metadata, inspect CSV datasets, craft valid identification
strategies, and run Python code to obtain causal effect estimates.

USER:
You are provided with a causal question, an inline metadata description,
and a CSV dataset (available as a file in your Python environment).

CAUSAL QUESTION:
{...}

METADATA ABOUT THE DATASET:
{...}

TASKS:
1) Read metadata to understand variables/context.
2) Load and analyze the CSV via Python to estimate the causal effect;
   report point estimate and standard error.
3) Fill out a JSON object with the identification specification,
   adhering to the provided schema.
4) Return the full Python script executed to produce the numbers.

FORMAT: Output a single JSON object with keys:
{ "identification_output": ..., "estimation_code": { "language": "python",
  "code": "...", "explanation": "..." } }
\end{lstlisting}
\end{tcolorbox}

\subsection{Aggregate Results}

Table~\ref{tab:llm-baseline-results} shows the aggregate performance of the baseline across all 174 queries.
The model correctly identifies the high-level strategy in 81.0\% of cases and the outcome variables in 96.0\% of cases.
However, performance drops sharply on more nuanced aspects of identification: causal quantity is correct in only 74.7\% of cases, and the overall identification specification is fully correct in only 37.9\% of cases.
This gap between high-level strategy recognition and full specification correctness is the central finding of our baseline evaluation, and it validates the design of CausalReasoningBenchmark: a single-score evaluation based on the final estimate would have obscured this important distinction.

\begin{table}[t]
\centering
\small
\caption{Aggregate evaluation of the GPT-5.3 baseline on all 174 queries. Identification metrics are exact-match or set-based checks against the gold specification; estimation metrics compare the returned effect and uncertainty to the gold solution. Values in brackets denote the interquartile range.}
\label{tab:llm-baseline-results}
\begin{tabular}{lr}
\toprule
\textbf{Metric} & \textbf{Value} \\
\midrule
\multicolumn{2}{l}{\emph{Identification Metrics}} \\
Strategy correct & 81.0\% (141/174) \\
Causal quantity correct & 74.7\% (130/174) \\
Treatments correct & 87.4\% (152/174) \\
Outcomes correct & 96.0\% (167/174) \\
Minimal controlling set included & 75.9\% (132/174) \\
Post-treatment set excluded & 94.3\% (164/174) \\
Controls correct & 70.7\% (123/174) \\
Strategy-specific fields correct & 90.8\% (158/174) \\
\textbf{Identification spec correct (all checks)} & \textbf{37.9\% (66/174)} \\
\midrule
\multicolumn{2}{l}{\emph{Estimation Metrics}} \\
Median absolute error $|\widehat{\tau}-\tau^\star|$  & 0.052 [0.005, 0.447] \\
Median percentage error  & 13.5\% \\
Median CI Jaccard overlap & 0.63 \\
CI Overlap & 89.1 \%\\
Estimate Within gold CI & 83.3 \% \\
Null Hypothesis Agreement & 79.3 \%\\
Opposite Direction Flag & 1.15 \%\\

\bottomrule
\end{tabular}
\end{table}

\subsection{Per-Strategy Breakdown}

Table~\ref{tab:per-strategy} provides a per-strategy breakdown of the baseline results. The model performs best on Regression Discontinuity queries, where the strategy is correct in 95.5\% of cases, though overall identification correctness is significantly lower. Second, Difference-in-Differences queries prove more challenging, particularly in specifying the correct time and group variables in panel data. Third, Regression Discontinuity queries show the largest gap between strategy-level correctness and full specification correctness, while Instrumental Variable queries also exhibit a substantial gap. 

\begin{table}[t]
\centering
\small
\caption{Per-strategy breakdown of the GPT-5.3 baseline. ``Strategy'' = fraction with correct strategy label; ``Full ID'' = fraction with fully correct identification specification; ``Med.\ \%Err'' = median percentage error on the effect estimate.}
\label{tab:per-strategy}
\begin{tabular}{lrrrr}
\toprule
\textbf{Design} & \textbf{\#Queries} & \textbf{Strategy (\%)} & \textbf{Full ID (\%)} & \textbf{Med.\ \%Err} \\
\midrule
Difference-in-Differences & 67 & 94.0 & 59.7 & 9.7 \\
Regression Discontinuity & 44 & 95.5 & 11.4 & 21.0 \\
Instrumental Variable & 22 & 86.4 & 36.4 & 50.1 \\
Conditional Exogeneity & 40 & 40.0 & 30.0 & 4.0 \\
RCT & 1 & 100.0 & 100.0 & 0.0 \\
\midrule
\textbf{Overall} & \textbf{174} & \textbf{81.0} & \textbf{37.9} & \textbf{13.5} \\
\bottomrule
\end{tabular}
\end{table}

%% ===== ANALYSIS AND QUALITATIVE ERROR ANALYSIS (merged) =====
%% This file is \input'd from main.tex and replaces the old
%% ``\subsection{Analysis}'' block and the standalone error-analysis section.

\subsection{Analysis}
\label{sec:error-analysis}

The baseline results reveal several important insights.
First, the large gap between strategy-level correctness (81.0\%) and full identification correctness (37.9\%) confirms that the bottleneck in automated causal reasoning lies not in recognizing the broad category of research design, but in specifying its detailed components. A single final-estimate score would not distinguish these failure modes.

Second, the estimation errors (median 13.5\% relative error, median Jaccard overlap of 0.63) are non-trivial but secondary to the identification errors.
In many cases, the model produces a reasonable estimate even when the identification specification is incorrect, because it may still use a plausible (but not gold-standard) approach.
This further underscores the importance of evaluating identification separately.

Third, the per-strategy breakdown reveals that different designs pose different challenges.
RDD queries are relatively easy to identify but may still have estimation errors due to bandwidth selection.
DiD queries require understanding temporal structure and group assignments.
IV queries demand the identification of a valid instrument---a task that requires deep domain knowledge.

\begin{figure}[t]
\centering
\includegraphics[width=\columnwidth]{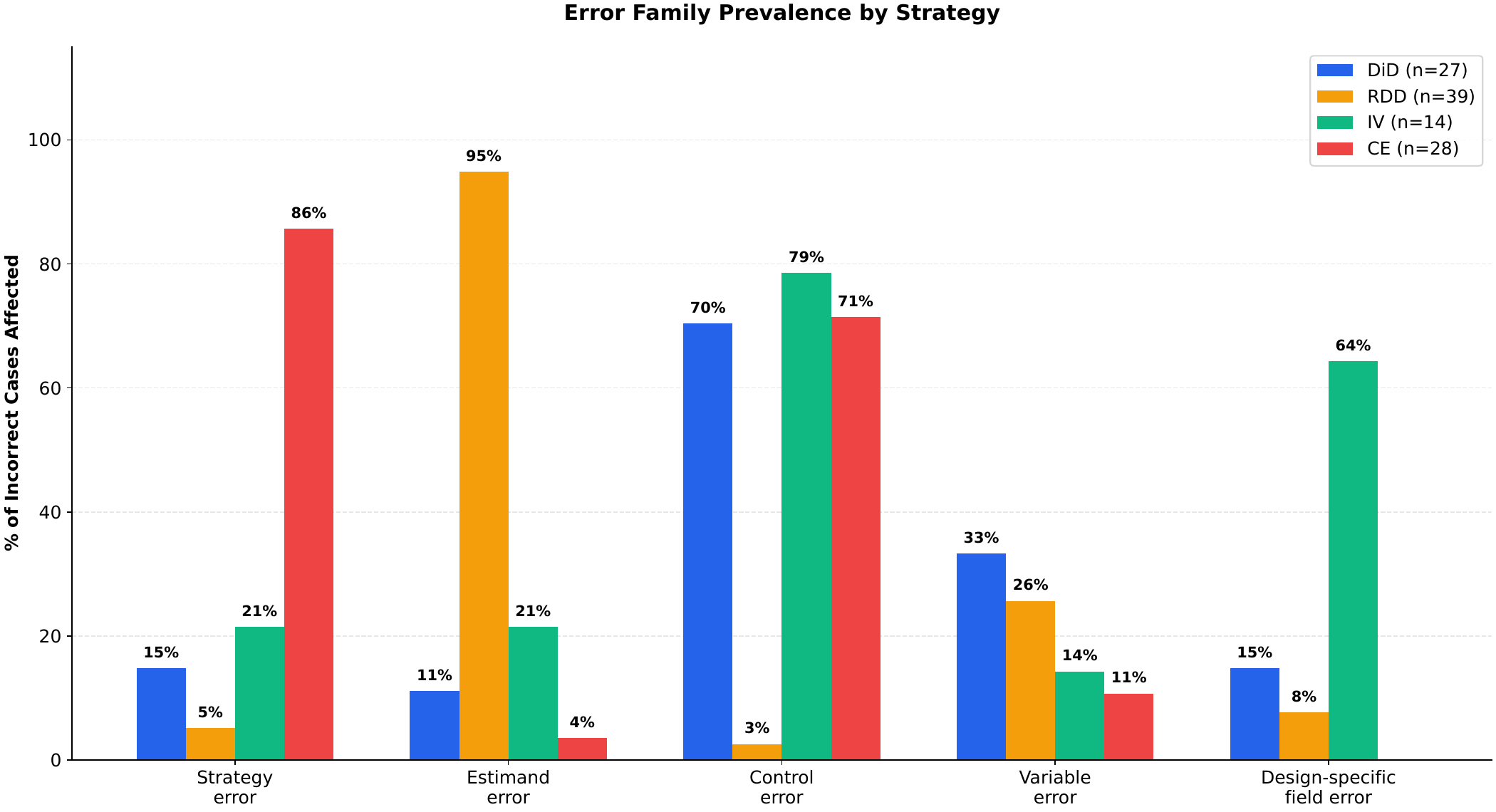}
\caption{Percentage of incorrect cases affected by each error family, grouped by gold strategy. Categories are \emph{non-exclusive}: a single case may exhibit multiple error types simultaneously (45\% of incorrect cases do). Each bar shows the fraction of that strategy's incorrect cases containing the given error family. RDD is dominated by estimand errors (95\%), CE by strategy errors (86\%), and DiD, IV, and CE all show high rates of control errors (70\%, 79\%, 71\%).}
\label{fig:strategy-error-decomp}
\end{figure}

To understand why the model fails, we categorize incorrect identification across the 108 failed cases.
Table~\ref{tab:error-taxonomy} summarizes the error types.
Figure~\ref{fig:strategy-error-decomp} further decomposes these errors by identification strategy, revealing that each design family has a distinct error profile.

\begin{table}[t]
\centering
\small
\caption{Taxonomy of identification errors across all 108 incorrect cases. A single case may contribute to multiple categories. Counts are grouped by error family.}
\label{tab:error-taxonomy}
\begin{tabular}{llr}
\toprule
\textbf{Error Family} & \textbf{Error Type} & \textbf{Count} \\
\midrule
Control variables & Missing required controls & 42 \\
 & Included bad / post-treatment controls & 10 \\
\midrule
Estimand & CATE $\to$ LATE & 28 \\
 & CLATE $\to$ LATE & 7 \\
 & Other estimand errors & 9 \\
\midrule
Strategy & CE $\to$ RCT & 24 \\
 & DiD $\to$ CE or IV $\to$ CE & 7 \\
 & Other strategy errors & 2 \\
\midrule
Variable specification & Wrong treatment variable & 22 \\
 & Wrong outcome variable & 7 \\
\midrule
Design-specific fields & Wrong IV-specific fields & 9 \\
 & Wrong RDD / DiD-specific fields & 7 \\
\bottomrule
\end{tabular}
\end{table}

\paragraph{Control variable errors.}
Control specification is the most error-prone component overall (42 cases with missing necessary controls, 10 with bad or post-treatment controls).
A representative omitted-control error occurs in the DiD questions on parliamentary gender quotas and public-health spending \citep{DiD_Clayton_Zetterberg_2018_2018}.
Both selected answers leave the control set empty, whereas the reference specification requires the lagged country characteristics \texttt{leftlag}, \texttt{demlag}, \texttt{lnODAlag}, \texttt{lnOillag}, and \texttt{lnTradelag}.
This is an omitted-adjustment error, not inclusion of the quota-induced change in women's legislative representation.

\paragraph{Missing the instrumental-variable design.}
Among the 22 IV queries, 9 cases exhibit errors in design-specific fields.
An example comes from a study of sibling composition and men's gender-role attitudes \citep{IV_Healy2013_2013}.
The paper instruments the share of female siblings (\texttt{share\_sis}) with whether the next-younger sibling is a sister (\texttt{closest}), adjusting for the number of siblings (\texttt{num\_sib}).
This design exploits the approximately random sex of the next-younger sibling, under the assumption that this sibling affects attitudes through overall sibling composition rather than an additional sibling-specific effect.
The model correctly identifies the treatment and outcome (\texttt{womens\_rights73}), but selects Conditional Exogeneity, reports ATE, and supplies no instrument, whereas the reference specification uses IV and labels the estimand LATE.
The error is therefore a failure to recover the paper's identifying variation, rather than simply an incorrect column name.

\paragraph{Strategy confusion: CE misidentified as RCT.}
The single most striking pattern is the model's tendency to classify observational studies requiring covariate adjustment as Randomized Controlled Trials (24 cases, all from the IHDP textbook subset studying the effect of specialist home visits on children's cognitive outcomes).
The metadata explicitly lists 25 baseline covariates---including birth weight, gestational age, maternal education, and neonatal health indicators---and a binary treatment indicator, but does \emph{not} state that treatment was randomly assigned.
The model appears misled by the word ``assignment'' in the query text, predicting RCT with an empty control set when the correct strategy is Conditional Exogeneity with all 25 covariates in the adjustment set.

\paragraph{Estimand confusion: CATE $\to$ LATE in RDD.}
In 28 of 44 RDD queries (64\%), the model correctly identifies the strategy but mislabels the estimand as LATE instead of CATE.
For instance, in a study of incumbency advantage in Japanese lower-house elections \citep{RDD_ariga_2015_2015}, the estimand is the treatment effect at the vote-margin cutoff---a conditional average treatment effect (CATE)---but the model labels it as LATE, conflating the RDD estimand with the complier effect from an IV design.

%% ===== SAMPLE QUERY =====
\section{Sample Query}
\label{sec:case}

To illustrate the task format, we present an example from \citet{coppock2016voting}.

\begin{tcolorbox}[colback=gray!5!white,colframe=gray!75!black,title=Example Causal Query]
\emph{``What is the causal effect of voting in the November 2007 municipal election on the probability of voting in the January 2008 primary, estimated for the subgroup of individuals whose November 2007 turnout is changed by the study's experimental encouragement?''}
\end{tcolorbox}

This query, along with the dataset and detailed metadata (excerpts below), is provided to the agent.

\begin{tcolorbox}[colback=gray!5!white,colframe=gray!75!black,title=First Few Rows of Provided CSV (excerpt):]
\begin{lstlisting}[basicstyle=\ttfamily\scriptsize,breaklines=true]
yob1,id,precinct,...,treatmen,hh,lastelec,voted,avhh0,color,hhsize,...
571129,00000026,506800006,...,control,489243,,,,,1,...
160626,00000030,705804003,...,control,934562,,,,,2,...
421017,00000066,638800041,...,control,880582,,,,,1,...
\end{lstlisting}
\end{tcolorbox}

\begin{tcolorbox}[colback=gray!5!white,colframe=gray!75!black,title=Partial Metadata:]
\begin{tabular}{ll}
\toprule
\textbf{Variable} & \textbf{Description} \\
\midrule
\texttt{treatmen} & Mailing-group label (e.g., ``control'', ``mail-arm A'') \\
\texttt{voted} & Binary indicator for voting (0/1) \\
\texttt{hhsize} & Household size (registered individuals) \\
... & ... \\
\bottomrule
\end{tabular}

\vspace{1em}
\noindent\textbf{Dataset Context:}
\begin{quote}
This dataset combines official voter-registration and voter-history records with household-level mailing assignments from a large mail program around municipal elections in 2007\ldots
\end{quote}
\end{tcolorbox}

\paragraph{Gold-Standard Identification.}
The correct identification for this query is an \textbf{Instrumental Variable} design.
The treatment is voting in the November 2007 election, the outcome is voting in the January 2008 primary, and the instrument is the randomized mailing assignment.
The causal quantity is \texttt{LATE}, because the IV design identifies the effect for compliers---those whose November 2007 turnout was changed by the mailing encouragement.
This example illustrates the level of reasoning required: the agent must recognize that the mailing assignment is a randomized encouragement (instrument) for the endogenous treatment (voting), and that the estimand is a LATE rather than an ATE.

%% ===== HOSTING =====
\section{Hosting and Maintenance}
\label{sec:hosting}

CausalReasoningBenchmark is publicly available on Hugging Face Datasets.\footnote{\url{https://huggingface.co/datasets/syrgkanislab/CausalReasoningBenchmark}} All associated code for evaluation, including the evaluator script and the gold-standard estimation scripts, is available in the same repository.

\paragraph{Licensing and Access.}
The benchmark metadata, evaluation code, and gold-standard identification specifications are released under the MIT license. The underlying datasets are redistributed under the terms of their original licenses; we provide attribution and licensing information for each dataset.

\paragraph{Maintenance Plan.}
We plan to update the benchmark periodically: adding queries as new reanalysis studies appear, incorporating additional designs (synthetic control, event studies), and revising metrics in response to feedback. We welcome contributions from the community.

%% ===== LIMITATIONS =====
\section{Limitations and Future Work}
\label{sec:limitations}

CausalReasoningBenchmark has several limitations that we plan to address in future work.

\paragraph{Domain Coverage.}
The research-paper subset is drawn entirely from political science, reflecting the availability of large-scale reanalysis studies in that field. While the textbook subset provides some cross-domain coverage, the benchmark would benefit from the inclusion of datasets from economics, epidemiology, and other fields where causal inference is central.

\paragraph{Strategy Coverage.}
The current benchmark focuses on five identification strategies.
Important designs such as synthetic control methods, event studies, and regression kink designs are not yet covered.
We plan to expand the strategy coverage in future releases.

\paragraph{Single Gold Standard.}
For each query, we provide a single gold-standard identification specification.
In practice, there may be multiple defensible identification strategies for a given dataset and question.
Future work could explore evaluation frameworks that accommodate multiple valid specifications.

\paragraph{Estimation Sensitivity.}
The gold-standard estimates are produced by specific estimation scripts. Different but equally valid estimation choices (e.g., different bandwidth selectors for RDD, different standard error clustering for DiD) could produce different estimates.

\paragraph{Scale.}
With 174 queries, CausalReasoningBenchmark is smaller than some existing benchmarks. However, we prioritize quality and depth of evaluation over quantity: each query requires a full identification specification, a python script to produce the causal estimate. We plan to expand the benchmark over time.

%% ===== CONCLUSION =====
\section{Conclusion}
\label{sec:conclusion}

CausalReasoningBenchmark provides a new, challenging, and realistic benchmark for evaluating automated causal reasoning systems.
By separating the evaluation of identification and estimation, it offers a more nuanced view of model capabilities than existing benchmarks.
Our baseline results demonstrate that state-of-the-art LLMs struggle with the detailed specification of causal research designs, even when they can correctly identify the broad design family. This finding highlights the need for more sophisticated reasoning capabilities in automated causal inference systems.
We hope that CausalReasoningBenchmark will help with the development of more robust and reliable AI systems for causal inference, and we welcome contributions from the research community.

\section{Acknowledgement}
\label{sec:ack}
Vasilis Syrgkanis, Ayush Sawarni and Jiyuan Tan were supported by NSF Award IIS-2337916.

\bibliographystyle{plainnat}
\bibliography{references,dataset_papers,dataset_textbooks}

\appendix
\section{Paper list and citations}
\label{sec:appendix-papers}
\small
\setlength{\LTpre}{0pt}
\setlength{\LTpost}{0pt}
\begin{longtable}{p{2.8cm}p{1.2cm}p{8.7cm}r}
\caption{Research papers included in the benchmark. Each row summarizes queries from one paper and design; papers used under multiple designs appear in separate rows.}\label{tab:paper-list}\\
\toprule
Paper & Design & Title & \#queries \\
\midrule
\endfirsthead
\toprule
Paper & Design & Title & \#queries \\
\midrule
\endhead
\bottomrule
\endfoot
\citep{DiD_Latura_Weeks_2023_2022} & DiD & Corporate Board Quotas and Gender Equality Policies in the Workplace & 2 \\
\citep{DiD_Ravanilla_2022_2022} & DiD & Deadly Populism: How Local Political Outsiders Drive Duterte's War on Drugs in the Philippines & 2 \\
\citep{DiD_Distelhorst_Locke_2018_2018} & DiD & Does Compliance Pay? Social Standards and Firm-Level Trade & 2 \\
\citep{DiD_Hainmueller_Hangartner_2019_2014} & DiD & Does Direct Democracy Hurt Immigrant Minorities? Evidence from Naturalization Decisions in Switzerland & 2 \\
\citep{DiD_Paglayan_2022_2022} & DiD & Education or Indoctrination? The Violent Origins of Public School Systems in an Era of State-Building & 2 \\
\citep{DiD_Zhang_etal_2021_2021} & DiD & Elite Cleavage and the Rise of Capitalism under Authoritarianism: A Tale of Two Provinces in China & 2 \\
\citep{DiD_Garfias_2019_2019} & DiD & Elite Coalitions, Limited Government, and Fiscal Capacity Development: Evidence from Bourbon Mexico & 2 \\
\citep{DiD_Blair_etal_2022_2022} & DiD & How Does Armed Conflict Shape Investment? Evidence from the Mining Sector & 2 \\
\citep{DiD_Esberg_Siegel_2023_2022} & DiD & How Exile Shapes Online Opposition: Evidence from Venezuela & 2 \\
\citep{DiD_Caughey_etal_2017_2017} & DiD & Incremental Democracy: The Policy Effects of Partisan Control of State Government & 2 \\
\citep{DiD_Magaloni_etal_2020_2020} & DiD & Killing in the Slums: Social Order, Criminal Governance, and Police Violence in Rio de Janeiro & 2 \\
\citep{DiD_Grumbach_2023_2022} & DiD & Laboratories of Democratic Backsliding & 2 \\
\citep{DiD_Dahlstrom_Holmgren_2023_2023} & DiD & Loyal Leaders, Affluent Agencies: The Budgetary Implications of Political Appointments in the Executive Branch & 2 \\
\citep{schafer2022making} & DiD & Making Unequal Democracy Work? The Effects of Income on Voter Turnout in Northern Italy & 2 \\
\citep{DiD_Eckhouse_2022_2021} & DiD & Metrics Management and Bureaucratic Accountability: Evidence from Policing & 2 \\
\citep{DiD_Kroeger_Silfa_2023_2023} & DiD & Motivated Corporate Political Action: Evidence from an SEC Experiment & 2 \\
\citep{DiD_Clarke_2020_2020} & DiD & Party Sub-Brands and American Party Factions & 1 \\
\citep{DiD_Kilborn_Vishwanath_2022_2021} & DiD & Public Money Talks Too: How Public Campaign Financing Degrades Representation & 2 \\
\citep{DiD_Clayton_Zetterberg_2018_2018} & DiD & Quota Shocks: Electoral Gender Quotas and Government Spending Priorities Worldwide & 2 \\
\citep{DiD_Grumbach_Sahn_2020_2019} & DiD & Race and Representation in Campaign Finance & 2 \\
\citep{schuit2017race} & DiD & Race, Representation, and the Voting Rights Act & 2 \\
\citep{DiD_Grumbach_Hill_2022} & DiD & Rock the Registration: Same Day Registration Increases Turnout of Young Voters & 2 \\
\citep{DiD_Liao_2023_2023} & DiD & The Effect of Firm Lobbying on High-Skilled Visa Adjudication & 2 \\
\citep{DiD_Fresh_2018_2018} & DiD & The Effect of the Voting Rights Act on Enfranchisement: Evidence from North Carolina & 2 \\
\citep{DiD_Trounstine_2020_2020} & DiD & The Geography of Inequality: How Land Use Regulation Produces Segregation & 2 \\
\citep{DiD_Hirano_etal_2022_2022} & DiD & The Growth of Campaign Advertising in the United States, 1880--1930 & 2 \\
\citep{DiD_Payson_2020_APSR_2020} & DiD & The Partisan Logic of City Mobilization: Evidence from State Lobbying Disclosures & 2 \\
\citep{DiD_Christensen_Garfias_2021_2021} & DiD & The Politics of Property Taxation: Fiscal Infrastructure and Electoral Incentives in Brazil & 2 \\
\citep{IV_Kuipers2022_2022} & DiD & The Representational Consequences of Municipal Civil Service Reform & 2 \\
\citep{DiD_Hankinson_Magazinnik_2023_2023} & DiD & The Supply-Equity Trade-Off: The Effect of Spatial Representation on the Local Housing Supply & 2 \\
\citep{DiD_Marsh_2023_2022} & DiD & Trauma and Turnout: The Political Consequences of Traumatic Events & 1 \\
\citep{DiD_Pierskalla_Sacks_2018_2018} & DiD & Unpaved Road Ahead: The Consequences of Election Cycles for Capital Expenditures & 2 \\
\citep{RDD_erikson_etal_2015_2015} & RDD & A Gubernatorial Helping Hand? How Governors Affect Presidential Elections & 2 \\
\citep{RDD_szakonyi_2018_2016} & RDD & Businesspeople in Elected Office: Identifying Private Benefits from Firm-Level Returns & 1 \\
\citep{RDD_palmer_schneer_2016a_2016} & RDD & Capitol Gains: The Returns to Elected Office from Corporate Board Directorships & 2 \\
\citep{RDD_carson_sievert_2017_2017} & RDD & Congressional Candidates in the Era of Party Ballots & 1 \\
\citep{RDD_schickler_etal_2009a_2010} & RDD & Congressional Parties and Civil Rights Politics from 1933 to 1972 & 4 \\
\citep{RDD_cavaille_marshall_2018_2024} & RDD & Education and Anti-Immigration Attitudes: Evidence from Compulsory Schooling Reforms across Western Europe & 2 \\
\citep{kim2019direct} & RDD & Direct Democracy and Women's Political Engagement & 1 \\
\citep{RDD_novaes_2018_2017} & RDD & Disloyal Brokers and Weak Parties & 1 \\
\citep{RDD_dahlgard_2018a_2002} & RDD & Trickle-Up Political Socialization: The Causal Effect on Turnout of Parenting a Newly Enfranchised Voter & 4 \\
\citep{RDD_folke_snyder_2012_2012} & RDD & Gubernatorial Midterm Slumps & 1 \\
\citep{DiD_Caughey_etal_2017_2017} & RDD & Incremental Democracy: The Policy Effects of Partisan Control of State Government & 1 \\
\citep{RDD_ariga_2015_2015} & RDD & Incumbency Disadvantage under Electoral Rules with Intraparty Competition: Evidence from Japan & 2 \\
\citep{RDD_eggers_spirling_2017_2017} & RDD & Incumbency Effects and the Strength of Party Preferences: Evidence from Multiparty Elections in the United Kingdom & 1 \\
\citep{coppock2016voting} & RDD & Is Voting Habit Forming? New Evidence from Experiments and Regression Discontinuities & 1 \\
\citep{RDD_holbein_hillygus_2016_2015} & RDD & Making Young Voters: The Impact of Preregistration on Youth Turnout & 1 \\
\citep{RDD_eggers_hainmueller_2009a_2009} & RDD & MPs for Sale? Returns to Office in Postwar British Politics & 1 \\
\citep{RDD_dbk_2018_2018} & RDD & Off-Cycle and Out of Office: Election Timing and the Incumbency Advantage & 2 \\
\citep{RDD_ferwerda_miller_2014_2014} & RDD & Political Devolution and Resistance to Foreign Rule: A Natural Experiment & 1 \\
\citep{RDD_broockman_ryan_2016_2015} & RDD & Preaching to the Choir: Americans Prefer Communicating to Copartisan Elected Officials & 1 \\
\citep{RDD_bohlken_2018_2018} & RDD & Targeting Ordinary Voters or Political Elites? Why Pork Is Distributed Along Partisan Lines in India & 1 \\
\citep{RDD_fouirnaies_hall_2014a_2014} & RDD & The Financial Incumbency Advantage: Causes and Consequences & 2 \\
\citep{RDD_klasnja_titiunik_2017_2017} & RDD & The Incumbency Curse: Weak Parties, Term Limits, and Unfulfilled Accountability & 1 \\
\citep{RDD_boas_etal_2014_2014} & RDD & The Spoils of Victory: Campaign Donations and Government Contracts in Brazil & 1 \\
\citep{RDD_hidalgo_nichter_2016_2015} & RDD & Voter Buying: Shaping the Electorate through Clientelism & 1 \\
\citep{RDD_hall_2015_2015} & RDD & What Happens When Extremists Win Primaries? & 2 \\
\citep{RDD_hall_thompson_2018_2018} & RDD & Who Punishes Extremist Nominees? Candidate Ideology and Turning Out the Base in US Elections & 1 \\
\citep{IV_Webster2022_2022} & IV & The Social Consequences of Political Anger & 1 \\
\citep{IV_Healy2013_2013} & IV & Childhood Socialization and Political Attitudes: Evidence from a Natural Experiment & 1 \\
\citep{IV_Flores2013_2016} & IV & The Foreign Policy Consequences of Trade: China's Commercial Relations with Africa and Latin America, 1992--2006 & 1 \\
\citep{dower2018collective} & IV & Collective Action and Representation in Autocracies: Evidence from Russia's Great Reforms & 2 \\
\citep{IV_Croke2016_2016} & IV & Deliberate Disengagement: How Education Can Decrease Political Participation in Electoral Authoritarian Regimes & 1 \\
\citep{IV_Delao2013_2012} & IV & Do Conditional Cash Transfers Affect Electoral Behavior? Evidence from a Randomized Experiment in Mexico & 1 \\
\citep{IV_Stokes2016_2015} & IV & Electoral Backlash against Climate Policy: A Natural Experiment on Retrospective Voting and Local Resistance to Public Policy & 1 \\
\citep{IV_Meredith2013_2013} & IV & Exploiting Friends-and-Neighbors to Estimate Coattail Effects & 1 \\
\citep{IV_Carnegie2017_2017} & IV & Foreign Aid, Human Rights, and Democracy Promotion: Evidence from a Natural Experiment & 1 \\
\citep{coppock2016voting} & IV & Is Voting Habit Forming? New Evidence from Experiments and Regression Discontinuities & 1 \\
\citep{IV_Gerber2010_2010} & IV & Party Affiliation, Partisanship, and Political Beliefs: A Field Experiment & 1 \\
\citep{IV_Lerman2017_2017} & IV & Personal Experience and Public Opinion: A Theory and Test of Conditional Policy Feedback & 1 \\
\citep{IV_Nellis2018_2017} & IV & Secular Party Rule and Religious Violence in Pakistan & 1 \\
\citep{IV_Rueda2017_2016} & IV & Small Aggregates, Big Manipulation: Vote Buying Enforcement and Collective Monitoring & 1 \\
\citep{IV_McClendon2014_2013} & IV & Social Esteem and Participation in Contentious Politics: A Field Experiment at an LGBT Pride Rally & 1 \\
\citep{IV_Lelkes2017_2015} & IV & The Hostile Audience: The Effect of Access to Broadband Internet on Partisan Affect & 1 \\
\citep{IV_Kuipers2022_2022} & IV & The Representational Consequences of Municipal Civil Service Reform & 1 \\
\citep{IV_Chong2019_2018} & IV & Urbanization Patterns, Information Diffusion, and Female Voting in Rural Paraguay & 1 \\
\end{longtable}

\end{document}